\documentclass[10pt,twocolumn,letterpaper]{article}
\usepackage{3dv}
\usepackage{times}
\usepackage{amsmath}
\usepackage{amssymb}
\usepackage{amsthm}
\usepackage{graphicx}
\usepackage{array}
\usepackage[pagebackref=true,breaklinks=true,colorlinks,bookmarks=false]{hyperref}
\usepackage{cleveref}
\usepackage{tikz,graphicx}
\usepackage{array}
\usepackage{adjustbox}
\usepackage{booktabs} 
\usepackage{float}
\usepackage{cuted}
\usepackage{subcaption}
\usepackage{soul}
\usepackage{multirow}

\usepackage{etoc}

\threedvfinalcopy

\newcommand{\methodColorActor}{NeuralDiff+C+A\xspace}
\newcommand{\methodColor}{NeuralDiff+C\xspace}
\newcommand{\methodActor}{NeuralDiff+A\xspace}
\newcommand{\method}{NeuralDiff\xspace}
\newcommand{\methodAblative}{NeRF-B\/F\xspace}

\newcommand{\benchmarkname}{EPIC-Diff\xspace}

\newcolumntype{L}[1]{>{\raggedright\let\newline\\\arraybackslash\hspace{0pt}}m{#1}}
\newcolumntype{C}[1]{>{\centering\let\newline\\\arraybackslash\hspace{0pt}}m{#1}}
\newcolumntype{R}[1]{>{\raggedleft\let\newline\\\arraybackslash\hspace{0pt}}m{#1}}

\makeatletter
\renewcommand{\paragraph}{%
  \@startsection{paragraph}{4}%
  {\z@}{0.25em}{-1em}%
  {\normalfont\normalsize\bfseries}%
}
\makeatother

\begin{document}

\title{\method: Segmenting 3D objects that move in egocentric videos}

\author{
Vadim Tschernezki$^{1,2}$ ~ ~ Diane Larlus$^2$ ~ ~ Andrea Vedaldi$^1$ \\
\centering
\begin{minipage}{.4\textwidth}
\centering
\small{~\\ $^1$Visual Geometry Group\\University of Oxford\\}
{\tt\small \{vadim,vedaldi\}@robots.ox.ac.uk} 
\end{minipage} 
\begin{minipage}{.4\textwidth}
\centering
\small{~\\~\\ $^2$NAVER LABS Europe\\}
{\tt\small diane.larlus@naverlabs.com} 
\end{minipage}
\vspace{-0.8cm}
}

\maketitle

\begin{strip}
\resizebox{\linewidth}{!}{
    \includegraphics[width=\linewidth]{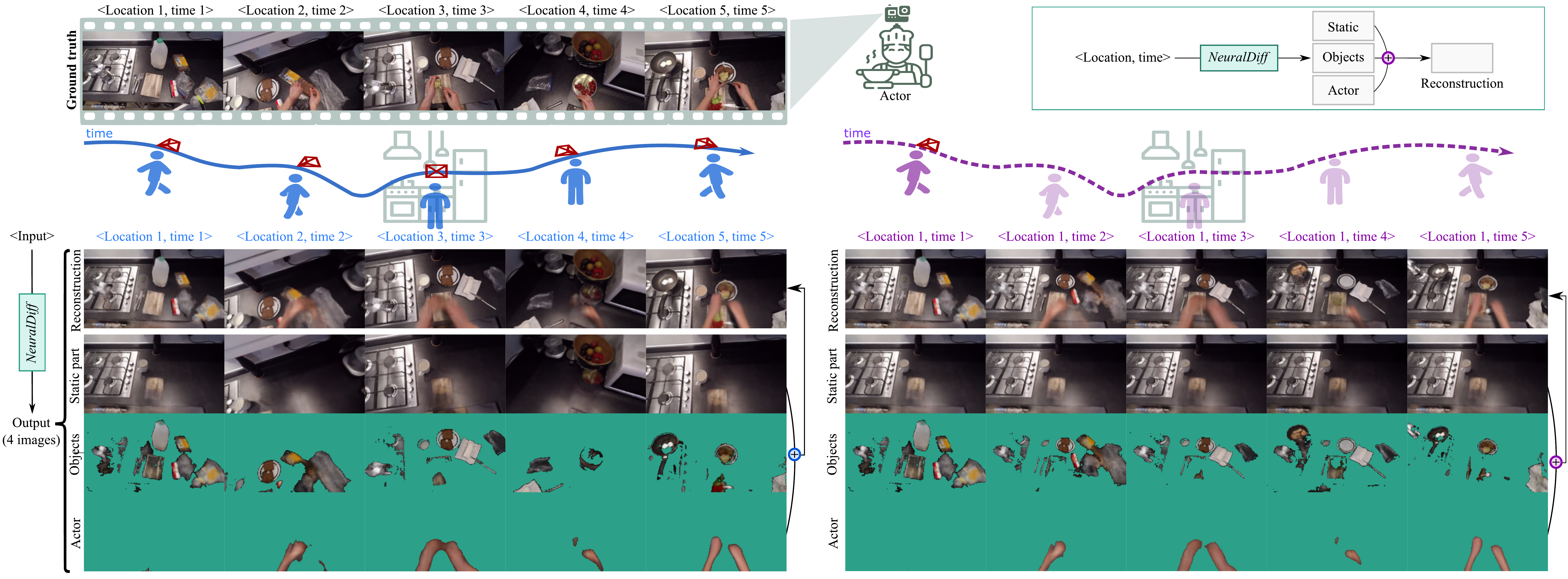}
    }
\captionof{figure}{
Given a video sequence captured from an egocentric viewpoint, we segment all the objects that the actor/observer interacts with.
We achieve this by means of a neural architecture, \method, that learns to decompose each frame into a static background and a dynamic foreground, comprising the manipulated objects, which seldomly move in the sequence, and the actor's body, which moves continually and heavily occludes the scene.
The neural network contains three streams that, via different inductive biases, reconstruct the background, the objects and the actor in 3D, and is thus able to render images and their segmentations also for viewpoints that do not exist in the original video sequence (see the left part of the figure, where the camera is assumed to remain at its initial position while the action unfolds).}
\label{fig:splash}
\end{strip}

\begin{abstract}
Given a raw video sequence taken from a freely-moving camera, we study the problem of decomposing the observed 3D scene into a static background and a dynamic foreground containing the objects that move within the scene.
This task is reminiscent of the classic background subtraction problem, but is significantly harder because all parts of the scene, static and dynamic, generate a large apparent motion due to the camera large viewpoint change and parallax.
In particular, we consider egocentric videos and further separate the dynamic component into objects and the actor that observes and moves them.
We achieve this factorization by reconstructing the video via a triple-stream neural rendering network that explains the different motions based on corresponding inductive biases.
We demonstrate that our method can successfully separate the different types of motion, outperforming recent neural rendering baselines at this task, and can accurately segment the moving objects. 
We do so by assessing the method empirically on challenging videos from the EPIC-KITCHENS dataset which we augment with appropriate annotations to create a new benchmark for the task of dynamic object segmentation on unconstrained video sequences, for complex 3D environments. 
Project page: {\small
\url{https://www.robots.ox.ac.uk/~vadim/neuraldiff/}}.
\end{abstract}

\section{Introduction}\label{s:introduction}

Given a video capturing a complex 3D scene, we consider the problem of segmenting the scene objects that move independently of the camera.
Motion is a powerful cue for discovering and learning visual objects in an unsupervised manner.
In fact, `detachability', namely the possibility of moving a body independently of the rest of the scene, is used by Gibson~\cite{gibson86the-ecological} to \emph{define} objects.
However, \emph{measuring} detachment given only raw visual observations as input is not an easy task.

If the video is taken from the viewpoint of a static camera, the problem of separating the static background from the moving foreground reduces to background subtraction.
However, classic background subtraction techniques are inapplicable if the camera undergoes a motion that induces significant parallax.
We may call this more challenging scenario \emph{wide-baseline background subtraction}.

To understand this concept, consider for example an egocentric video 
of a person cooking.
This actor intervenes in the scene by moving (and transforming) objects.
However, egomotion is the dominant effect: by comparison, objects move only sporadically, and in a way that is hardly distinguishable from the much larger apparent motion induced by the viewpoint change.
Extracting the moving objects automatically is thus very difficult, and essentially impossible for traditional background subtraction techniques.

One may use motion segmentation techniques to separate a scene in different motion components.
However, these techniques generally require correspondences (\eg, optical flow), they reason locally, across a handful of frames, and usually avoid explicit 3D reasoning.
In short, they are of difficult applicability to video sequences such as the ones in~\Cref{fig:splash}, comprising many small rigid objects that move only occasionally throughout a long sequence.

In this paper, we propose to leverage recent progress in neural rendering techniques~\cite{mildenhall20nerf:} 
to develop a motion \emph{analysis} tool
to achieve the desired segmentation.
We build on the ability of neural rendering to reconstruct accurately the appearance of a rigid 3D scene under a variable viewpoint, without requiring dense correspondences.
Given the reconstruction of the background, it is then possible to measure the more subtle appearance `differences' induced by the objects that move \emph{independently} of the camera.

We further note that the 3D objects manipulated in the video also contain significant structure.
Specifically, they move in `bursts', changing their state as they are manipulated, but remaining otherwise rigidly attached to the background.
We thus extend the neural renderer to also reconstruct the object appearance using a slowly-varying time encoding for them.
In fact, we go one step further and introduce a third neural rendering stream that captures the actor observing and moving the objects.
The intuition is that the actor moves continually, in a way that occludes the scene, with a motion linked to the camera and not to the scene (being the observer), leading to significantly different dynamics compared to the background and foreground objects.

\begin{figure*}[t!]
    \centering
    \includegraphics[width=0.85\linewidth]{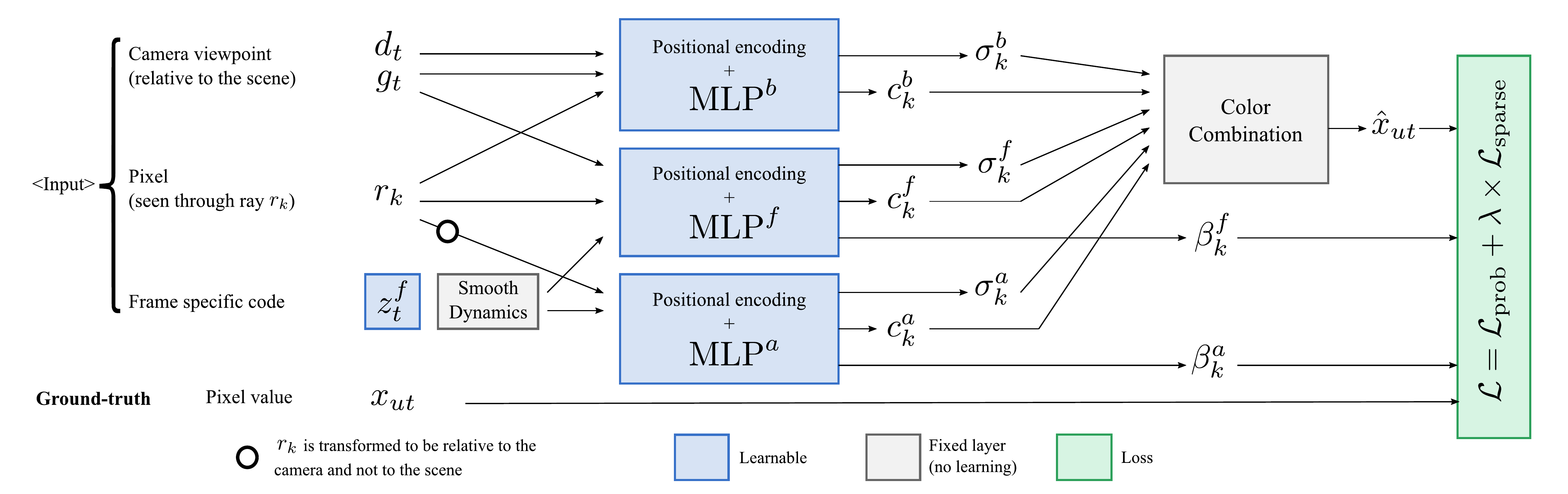}
    \vspace{-0.2cm}
    \caption{\textbf{Overview of \method}. Given only the camera viewpoint $g_t$ and a frame specific code $z_t^f$ (learned latent variable), our three stream architecture learns to predict the color value of pixel $x_{ut}$ by combining information coming from the static model of the background, and from two dynamic components, one for the foreground objects, and one for the actor. The parameters of this model are learned using a probabilistic loss $\mathcal{L}$.}
    \label{fig:architecture}
\end{figure*}

Our technical contribution is thus a three-stream neural rendering architecture, where the streams model respectively i) the static background, ii) the dynamic foreground objects, and iii) the actor. Those are then composed to explain the video as a whole (see \Cref{fig:architecture}).
We design the streams differently, in order to incorporate inductive biases that match the statistics of each layer (background, foreground, actor).
These inductive biases depart from previous neural rendering models because of the structure of the foreground, which is composed of several objects being manipulated at different intervals, and of the actor, a deformable body attached to the camera and not to the background.

The resulting analysis-via-synthesis method shows that neural rendering techniques are not only useful for synthesis, but also for analysis.
In particular, we are the first to demonstrate the effectiveness of these techniques in interpreting challenging egocentric videos, providing cues for the extraction of detached objects in scenes with a complex 3D structure and dynamics.

We focus our empirical evaluation on egocentric videos because, with the emergence of AR, they are becoming increasingly popular and have the advantage of showing the interaction of actors with their environment.
We expect this kind of videos to provide an enormous wealth of information for computer vision, particularly with the recent introduction of Ego4D~\cite{grauman21ego4d}.
They are also particularly challenging to process, providing an excellent test scenario for this class of algorithms.

For evaluation, we augment the EPIC-KITCHENS dataset~\cite{damen18epic} and manually segment all objects that move at some point during the scene, \ie that are thus detached.
With these annotations, we can assess neural renderers not only in terms of new view synthesis quality, but also, and more to the point, in terms of their ability to separate videos in the various dynamic components.
With this, we also define a new \emph{benchmark} for measuring progress in the challenging task of dynamic object segmentation in complex videos, thus inviting further research in the area.

Using this data, which we call \benchmarkname, we show that our model outperforms the direct application of existing neural rendering approaches such as NeRF~\cite{mildenhall20nerf:} or NeRF-W~\cite{martinbrualla2020nerfw} in the wide-baseline background subtraction problem.

\section{Related work}\label{s:related}

Although unique, our research problem relates to several existing research directions that we mention below.

\paragraph{Background subtraction.} Background subtraction techniques have been used for detecting moving objects in video sequences (see survey~\cite{bouwmans2014traditional}). They typically assume that only the foreground moves, and many methods rely on a background initialization step which assumes that the first few frames give a good estimate of the background's appearance. The background model can be updated during the sequence, but these methods fail when the background varies as much as in the egocentric videos that we consider~\cite{kalsotra2019comprehensive}.  

\paragraph{Motion segmentation.}
Closely related to background subtraction, motion segmentation is the more general task of decomposing a video into individually moving objects \cite{08motionseg, 21motionseg, OB14b}. 
These techniques generally rely on optical flow, which is subject to ambiguities \cite{cvpr20motionseg}, and reasons only locally without a 3D representation.
Occlusions are one of the main challenges \cite{cvpr21motionseg}.
Many methods fail when dynamic objects temporarily remain static, even for a few frames \cite{08motionseg, 21motionseg}.
These issues prevent applying standard motion segmentation methods to egocentric videos (where objects rarely move and the actor heavily occludes the scene).

\paragraph{Discovering and segmenting objects in videos.}
This long-standing problem is related to background subtraction and motion segmentation \cite{bideau2016s, papazoglou2013fast, tokmakov2017learning, jain2017fusionseg, xie2019object}.
For example, one can use a probabilistic model that acts upon optical flow to segment moving objects from the background~\cite{bideau2016s}. In~\cite{sundaram2010dense, brox2010object}, pixel trajectories and spectral clustering are combined to produce motion segments. 
The method of~\cite{matzen14scene} uses image collections to reconstruct urban scenes and to discover their dynamic elements such as billboards or street art, by clustering 3D points in space and time. 
Recent work revisits classical motion segmentation techniques from a data-driven perspective~\cite{Yang21a, wang2019zero, xie2019object}, \eg using physical motion cues to learn 3D representations~\cite{tenenbaum00a-global} or learning a factored scene representation with neural rendering~\cite{yuan2021star}. 
Our work departs from these approaches and learns a holistic representation capable of handling occlusions and sporadically moving objects, without the need for multi-view videos as in~\cite{yuan2021star}. We approach the problem on complex real-world data as opposed to clean synthetic ones as in \cite{tenenbaum00a-global}.

\paragraph{Neural rendering.} Neural rendering is a way of synthesizing novel views using a neural architecture and classic volume rendering techniques.
It was introduced as Neural Radiance Fields (NeRF) in \cite{mildenhall20nerf:} for static scenes. By default, it poorly handles dynamic scenes and occlusions.
It is extended in \cite{martinbrualla2020nerfw} as NeRF-W for the case of unconstrained photo collections to deal with transient objects occluding the static scene. Another research direction is studied in \cite{xie2021fignerf, stelzner2021decomposing} to model the 3D geometry of one or several static objects.
Recent work has also focused on modeling dynamic scenes, mostly with monocular videos as input \cite{park2020nerfies,li21neural, pumarola2021d, chen2021animatable, gao2021dynamic, tretschk2021nonrigid}.
Most approaches~\cite{park2020nerfies,pumarola2021d,chen2021animatable,peng2021animatable} combine a canonical model of the object with a deformation network, or warp space~\cite{tretschk2021nonrigid}, still starting from a canonical volume. 
Closer to our work, \cite{li21neural} and \cite{gao2021dynamic} combine a static NeRF model and a dynamic one.
Yet, none of those methods explicitly tackle the segmentation of 3D objects in such long and challenging video sequences.

\section{Method}
\label{sec:method}

Given a video sequence $x$, we wish to extract a corresponding mask $m$ that separates the foreground objects from the background.
We define as background the part of the scene that remains static throughout the entire video, generating an apparent motion only due to the camera viewpoint change.
We define as foreground any object that moves independently of the camera in at least one frame.

Traditional background subtraction techniques solve a similar problem by predicting the appearance of each video frame \emph{as if} the foreground objects were removed; given this prediction, the foreground objects can be segmented by taking the difference between the measured and predicted images.
Predicting the appearance of the background under an occluding foreground object is relatively easy for a static camera, where correspondences with other video frames where the object is not present can be trivially established.
However, the prediction is much more challenging if the viewpoint is also allowed to change.

In order to solve this problem, we build on recent neural rendering techniques such as NeRF~\cite{mildenhall20nerf:} that can predict effectively the appearance of a static object, in our case the rigid background, under a variable viewpoint.
In fact, we suggest that the foreground objects, which change over time, can \emph{also} be captured via a (distinct) neural rendering function, further promoting separation of background and foreground.
Next, we first discuss neural rendering in general, and then introduce our model.

\subsection{Neural rendering}
\label{sec:basic_nerf}

We base our method on NeRF~\cite{mildenhall20nerf:}, which we summarize here.
A video $x$ is a collection $(x_t)_{t\in[0,\dots,T-1]}$ of $T$ video frames, each of which is an RGB image $x_t \in \mathbb{R}^{3\times H\times W}$.
The video frames are a function $x_t=h(B,F_t,g_t)$ of the static background $B$, the variable foreground $F_t$ and a moving camera $g_t \in SE(3)$, where $SE(3)$ is the group of Euclidean transformations.
The motion $g_t$ is assumed to be known, estimated using an off-the-shelf SfM algorithm such as  COLMAP~\cite{schoenberger16sfm,schoenberger16mvs}.
The background and foreground components comprise the shape and reflectance of the 3D surfaces in the scene as well as the illumination.

Rather than attempting to invert $h$ to recover $B$ and $F_t$, which amounts to inverse rendering, neural rendering \emph{learns} the mapping $h$ directly, as a neural network $f$, $h(B,F_t,g_t)\approx f(g_t,t)$, providing time and viewpoint to reconstruct the corresponding video frame $x_t$.
By a careful design of the function $f$, the learning process can induce a factorization of viewpoint $g$ and time $t$, thus generalizing $f$ to new (unobserved) viewpoints.
NeRF additionally assumes that the scene is static, meaning that the variable foreground $F_t$ is empty, so the function simplifies to $f(g_t)$.
The model $f(g_t)$ is further endowed with a specific structure, which is key to successful learning~\cite{zhang20nerf:}.
Specifically, the color $x_{ut} \in \mathbb{R}^3$ of pixel $u \in \Omega = \{0,\dots,H-1\}\times \{0,\dots,W-1\}$ is obtained by a volumetric sampling process that simulates ray casting.
One `shoots' a ray
$
r_k = \ell_k K^{-1}(u)
$
along the viewing direction of pixel $u$, where 
$
K : \mathbb{R}^2\times \{1\} \rightarrow \Omega
$ 
is the camera calibration function and $\ell_0 < \cdots < \ell_{M+1} \in \mathbb{R}_+$ are the sampled depths.

The pixel color is obtained by averaging the color of the 3D points $g_t r_k$ along the ray, weighed by the probability that a photon emanates from the point and reaches the camera.
A neural network,
\newcommand{\MLP}{\operatorname{MLP}}
$
(\sigma_k^b,c_k^b) = \MLP^b(g_t r_k, d_t),
$
estimates the density $\sigma_k^b \in \mathbb{R}_+$ and the color $c_k^b\in\mathbb{R}^3$ of each point $g_t r_k$, where the superscript $b$ denotes the fact that the quantities refer to the background `material' and $d_t$ is the unit-norm viewing direction.

The probability that a photon is transmitted while traveling through the ray segment $(r_k, r_{k+1})$ is defined to be
$
T^b_k = e^{-\delta_k \sigma^b_k}
$
where the quantity $\delta_k = |r_{k+1}-r_k|$ is the length of the segment.
This definition is consistent with the fact that the probability of transmission across several segments is the product of the individual transmission probabilities.
We can thus write the color of pixel $u$ as:
\begin{equation}
x_{ut}
=
f_u(g_t)
=
\sum_{k=0}^M
v_k\,
(1 - T^b_k)
\,
c^b_k,
~~~~
v_k = \prod_{q=0}^{k-1} T^b_q.
\end{equation}
The network is trained by minimizing the reconstruction error $\|x - f(g_t)\|$, thus fitting a single video at a time.

\subsection{Dynamic components}\label{sec:foreground_modelling}

The method discussed above assumes that the scene is rigid.
In our case, reconstructing the scene is more complex due to the variable foreground $F_t$.
In order to capture this dependency, on top of the background MLP ($\MLP^b$), we introduce a foreground-specific MLP,
$
(\sigma^f_k,c^f_k,\beta^f_k) = \MLP^f(g_t r_k, z^f_t).
$
It produces a `foreground' occupancy $\sigma^f$ and color $c^f$. 
Additionally, it predicts an uncertainty score $\beta^f_k$ whose role is clarified in \Cref{sec:uncertainty}.
We also introduce a dependency on a frame-specific code $z^f_t\in\mathbb{R}^D$, capturing the properties of the foreground that change over time.

The color $x_{ut}$ of a pixel $u$ is obtained by composition 
of multiple materials $\mathcal{S}$ (\eg the background and the foreground, so $\mathcal{S}=\{b,f\}$):

\begin{multline}
x_{ut}
=
f_u(g_t,z_t)
=
\sum_{k=0}^M
v_k
\left( \sum_{p\in \mathcal{S}} w^p(T_k) c^p_k \right)
, \\
\text{where}\qquad v_k = \prod_{q=0}^{k-1} \prod_{p \in \mathcal{S}} T_q^p
\label{e:comp}
\end{multline}
The factor $v_k$ requires a photon to be transmitted from the camera to point $r_k$ through the different materials (hence the transmission probabilities are multiplied).
The weights $w^p(T_k)$ mix the colors of the materials proportionally to their density.
Following NeRF-W~\cite{martinbrualla2020nerfw}, we can simply set
\begin{equation}\label{e:mix1}
  w^p(T_k) = 1 - T_k^p \in [0,1].
\end{equation}

\paragraph{Smooth dynamics.}

The model (MLP) weights and the frame-specific parameters $z_t$ can be optimized by minimizing the loss across all input frames
$$
\min_{f,z_1,\dots,z_T}
\frac{1}{T|\Omega|}\sum_{t=1}^T\sum_{u\in\Omega}
\| x_t - f(g_t,z_t) \|^2
$$
in an auto-decoder fashion.
However, the foreground, while dynamic, does not change arbitrarily from frame to frame.
In particular, most foreground objects are in most frames rigidly attached to the background.
Because of this, the dependency on independent frame-specific codes $z_t$ makes little sense;
we replace it with a low-rank expansion of the trajectory of states, setting $z_t = B(t) \Gamma $ where $B(t) \in \mathbb{R}^ {P}$ is a simple handcrafted (fixed) basis and the motion $\Gamma \in \mathbb{R}^{P\times D}$ are coefficients such that $P \ll T$.
We take in particular $B(t) = [1, t, \sin 2\pi t, \cos 2\pi t, \sin 4 \pi t, \cos 4\pi t,\dots ]$ to be a deterministic harmonic coding of time (meaning that $z_t$ varies slowly over time.)

The method described above with a static part (as defined in \Cref{sec:basic_nerf}) and a dynamic part describing the objects ($\mathcal{S} = \{b,f\}$) is the basic version of our proposed approach. We refer to it as \texttt{\method} in what follows.

\paragraph{Improved geometry: capturing the actor.}
In egocentric videos, we further distinguish the foreground objects manipulated by the actor/observer, which moves sporadically, from the actor's body, which moves continually.
To model the latter, we consider a third MLP tasked with capturing parts of the actor's body that appear in the frames.
Formally, the actor MLP is similar to the foreground MLP\@:
$
(\sigma^a_k,c^a_k,\beta^a_k) = \MLP^a(r_k, z^a_t),
$
with the key difference that the 3D point $r_k$ is expressed relative to the camera (v.s. $g_tr_k$ which is expressed relative to the world).
This is due to the fact that the camera is anchored to the actor's body, which therefore shows a reduced variability in the reference frame of the camera.
By contrast, the background is invariant if expressed in the reference frame of the world; the same is true for the foreground objects when they are not manipulated, which is true most of the time.
This inductive bias helps factoring the different materials. Here $\mathcal{S} = \{b,f,a\}$.

We refer to this new flavor as \texttt{\methodActor}.

\paragraph{Improved color mixing.}

Eq.(\ref{e:mix1}), used in prior work to mix colors from different model components, cannot be justified probabilistically as it amounts to summing non-exclusive probabilities (nothing in the model prevents two or more materials to have non-zero density at a given point).

A principled mixing model is obtained by decomposing the segment $\delta_k$ in $Pn$ sub-segments, alternating between the $P$ different materials ($P=|\mathcal{S}|$, \eg $P=3$ if the background, foreground and actor are considered).
In the limit, we can show that the probability that the photon is absorbed in a subsegment of material $p$ is given by:
\begin{equation}\label{e:mix2}
w^p(T_k)
= \frac{\sigma^p_k}{\sum_{q=1}^P \sigma^q_k}\left(1 - \prod_{q=1}^P T^q_k\right).
\end{equation}
The second factor in parenthesis is, evidently, the probability that the photon is absorbed by any of the materials.
The first factor, which involves the densities rather than the probabilities, is the probability that a given material $p$ is responsible for the absorption.
Note that, differently from~\cref{e:mix1}, this definition yields
$
\sum_{p=1}^P w^p(T_k) = 1 - \prod_{q=1}^P T^q_k = 1 - v_k / v_{k-1}
$
which is consistent with the definition of transmission probability $v_k$.
A proof of~\cref{e:mix2} is available in the supplementary material. 

The \method model can thus be improved by taking into account this more principled way of producing pixel colors in the reconstruction. We refer to this improved version as \texttt{\methodColor}. 

Note that the two proposed improvements are complementary and we refer to the method enhanced by both as \texttt{\methodColorActor}. 

\subsection{Uncertainty and regularization}
\label{sec:uncertainty}

\paragraph{Uncertainty.}

The MLPs also predict scalars $\beta^p_k \geq 0$ (where, in practice, $\beta^b_k=0$ for the background).
These are used to express the uncertainty of the color associated to each 3D point $r_k$ for each material $p$ as pseudo-standard deviations (StDs).
Following~\cite{martinbrualla2020nerfw}, the StD of the rendered color $x_{ut}$ is just the sum of the StDs $\beta_{ut} = \sum_{p}\beta^p_{ut}$, where $\beta^p_{ut}$ is obtained via \cref{e:comp} by `rendering' the StDs $\beta^p_k$ of each 3D material point (it suffices to replace $\beta^p_k$ for $c^p_k$ in~\cref{e:comp}).
The StD $\beta_{ut}$ is used in a Gaussian observation loss as a form of self-calibrated aleatoric uncertainty~\cite{novotny17learning,kendall17what}:
\begin{equation}\label{e:loss1}
  \mathcal{L}_\text{prob}(f,z_t|x_t,g_t,u)
  =
  \frac{\| x_{ut} - f_u(g_t, z_t) \|^2}{2\beta_{ut}^2}
  + \log \beta_{ut}^2.
\end{equation}

\paragraph{Sparsity.}

We follow NeRF-W~\cite{martinbrualla2020nerfw} and further penalize the occupancy of the foreground and actor components using an $L^1$ penalty:
$$
\mathcal{L}_\text{sparse}(f,z_t|x_t,g_t,u)
= \sum_{p=1}^P\sum_{k=0}^M \sigma^p_k.
$$
This is the $L^1$ norm of the ray occupancies, which encourages the foreground occupancy to be sparse.

\paragraph{Training loss.}
Finally, the model is trained using the loss 
$
\mathcal{L} = \mathcal{L}_\text{prob} + \lambda \mathcal{L}_\text{sparse}
$
where $\lambda > 0$ is a weight set to 0.01.

\subsection{From \method to a scene segmentation}

Our approach is trained for a reconstruction task, but our primary goal is wide-baseline background subtraction, so we need to extract masks that assign a given pixel to the background, foreground, or actor layers.
We do so by constructing three indicator channels and we use \cref{e:comp} to `render' them as a mask $m_{ut} \in \mathbb{R}^3$ --- in other words, we simply associate pseudo-colors $(1,0,0)$, $(0,1,0)$ and $(0,0,1)$ to all 3D points of background, foreground and actor, respectively, and use~\cref{e:comp} to obtain $m_{ut}$.
\section{\benchmarkname benchmark}\label{s:dataset}

Our goal is to identify any object which moves independently of the camera in a video sequence.
We create a suitable benchmark for this task by augmenting the well-known EPIC-KITCHENS dataset~\cite{Damen2020RESCALING} with new annotations.
EPIC-KITCHENS is an egocentric dataset, with 100 hours of recording, 20M frames, and 90'000 actions performed. 

\paragraph{Data selection.}
%
We selected 10 video sequences, each lasts 14 minutes on average. Then we uniformly 
extracted 1'000 frames from each, and preprocessed them with COLMAP~\cite{schoenberger16sfm} to obtain the camera calibration and extrinsics (motion).
We refer to each video sequence as a \textit{scene}. 
To select the scenes, we considered two constraints.
First, the videos should contain a diversity of viewpoints and manipulated objects.
Second, COLMAP must successfully reconstruct the scene and register at least 600 frames.
We only retain the frames where COLMAP succeeds, which results in an average of 900 frames per scene.

\begin{figure}[t!]
    \centering
        \includegraphics[width=0.24\linewidth]{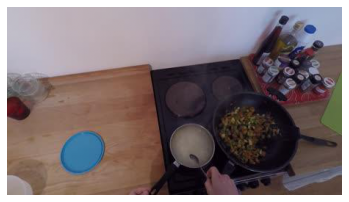} 
        \includegraphics[width=0.24\linewidth]{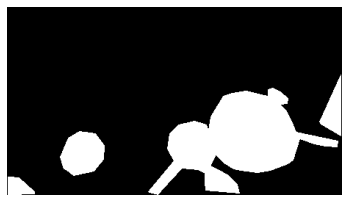} 
        \includegraphics[width=0.24\linewidth]{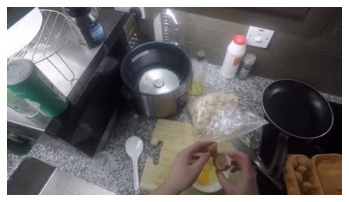} 
        \includegraphics[width=0.24\linewidth]{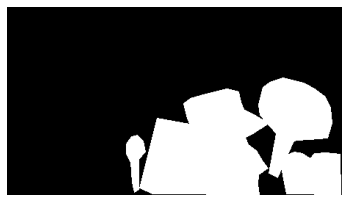}
    \vspace{-0.5em}
    \caption{Two frames from \benchmarkname and their corresponding manual foreground/background segmentation masks.}
    \label{fig:data}
\end{figure}

\paragraph{Data annotation.}
Since our algorithms are unsupervised, we do not need to collect extensive data annotations.
We uniformly 
hold out on average 56 frames for validation (for setting parameters) and 56 for testing. We annotated the latter with segmentation masks to assess background/foreground segmentation.
These frames are \emph{not} used to train the model, so they can also be used to assess new-view synthesis.
A frame annotation consists of a pixel-level binary segmentation mask, where the foreground contains any pixel that belongs to an object that is observed moving \emph{at any point in time} during the video.
Note that the fact that an object is marked as foreground in a frame does not mean it moves in \emph{that} frame; it only means that it moves at least once in the scene.
Based on this definition, the foreground mask covers both foreground objects and the actor. 
We obtain about 560 manual image-level segmentation masks.
Examples of these masks are given in~\Cref{fig:data}.

\paragraph{Evaluation.}

We evaluate the wide-baseline background subtraction task.
For this, we use standard segmentation metrics:
for each test frame, we sort all pixels based on their foreground score as defined above, and use the ground truth binary mask to compute average precision (AP).
We then report mAP by averaging across frames and scenes.
We also evaluate the new-view synthesis quality by tasking our model with synthesizing each of the test views, and we measure PSNR.
Specifically, we report PSNR for different parts of the scene using the ground-truth segmentation masks: the whole image, the background and the foreground regions.

\section{Experiments}\label{s:experiments}

\subsection{Experimental settings}\label{sec:exp_protocol}

\paragraph{Implementation details.}

All reported experiments are based on a PyTorch implementation of NeRF~\cite{mildenhall20nerf:} extended with several NeRF-W components~\cite{martinbrualla2020nerfw}. 
The architecture combines several interconnected MLPs to model density, color and uncertainty of background, foreground and actor.
Their architecture's details are given in the supplementary material.
We reuse the same motion coefficients $z^f_k=z^a_k=z_k\in\mathbb{R}^{17}$ for both foreground and actor (see~\cref{sec:foreground_modelling}).
Positional encoding is used to encode position, motion coefficients, and viewing direction using respectively 10, 10 and 4 frequencies. 
All models are trained using the Adam optimizer with an initial learning rate of $5 \times 10^{-4}$ using a cosine annealing schedule \cite{loshchilov2017sgdr}.
For more details regarding this implementation and differences to NeRF and NeRF-W, see the supplementary material.

\newcolumntype{C}{>{\centering\arraybackslash}p{2.5cm}}

\begin{figure*}[t!]
    \centering
    \vspace{-1em}
    \small
    \begin{tabular}{cCCCCC}
        ~~~~~~~~ & ~~~~Ground-Truth & NeRF & NeRF-W & \method & \!\!\methodColorActor \\
    \end{tabular}
    \vspace{-1em}
    \includegraphics[width=0.85\linewidth]{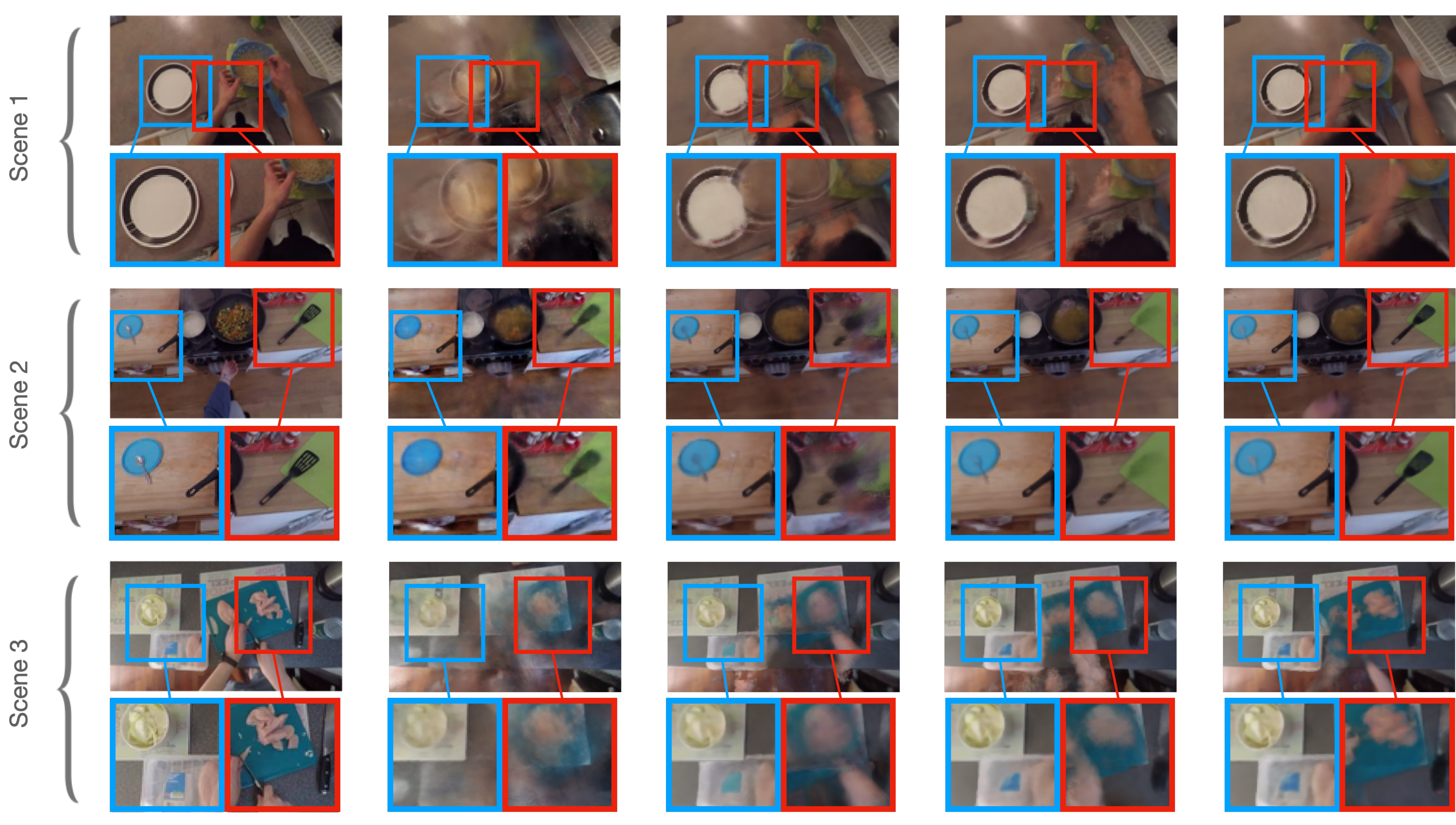}
    \caption{For three different scenes, reconstruction produced by NeRF~\cite{mildenhall20nerf:}, NeRF-W~\cite{martinbrualla2020nerfw}, \method and \methodColorActor. Our method has less ghosting artifacts, captures most moving objects, and shows more details.}
    \label{fig:quali_comparisons}
\end{figure*}


\paragraph{Baselines.}

We compare against the following baselines:

(1) \textbf{NeRF}~\cite{mildenhall20nerf:} uses a single stream.
As it has been designed to only capture the static part of a scene (background), we use the per-pixel prediction error as a pseudo foreground score (the larger the error, the more likely for the pixel to belong to the foreground, because NeRF cannot explain it with its static model).

(2) \textbf{\methodAblative} trains two parallel NeRF models, one for the background (B) and one for the foreground (F).
The F stream is further conditioned on time, but passing a positional-encoded version of the time variable $t$ (frame number) as input, instead of the learnable dynamic frame encoding of our method.

(3) \textbf{NeRF-W}~\cite{martinbrualla2020nerfw} also contains two interlinked background and foreground streams. Because it was initially proposed for image collections and not videos, by default the foreground parameters $z^f_t$ are learned independently for each frame.\footnote{In NeRF-W, the foreground is designed to capture the transient part of the scene that should be ignored (\eg persons occluding a landmark).}
This design limits the applicability of NeRF-W, as it is unable to render novel views for frames not available in the training set.
We redesign NeRF-W as a baseline for our task by adjusting the frame specific code such that it produces the foreground using the code of its closest neighboring frame from the training set.
More precisely, given a test frame $t$, we set the code for this test view $z^f_t := z^f_j$ where $j$ is the closest frame to $t$ out of all training frames $\mathcal{I}$ (\ie $j = \arg\!\min_{i}(\{|i - t|:i \in \mathcal{I}\})$.
We follow a similar strategy for the frame specific code for appearance (responsible for capturing the photometric variations).


\begin{table}[t!]
  \centering
  \small
  \begin{tabular}{lccccc}
    \toprule
    Method & mAP & PSNR & $\text{PSNR}_b$ & $\text{PSNR}_f$ \\
    \midrule
    NeRF~\cite{mildenhall20nerf:} & 47.8 & 20.9  & 22.8  & 17.6 \\
    NeRF-W~\cite{martinbrualla2020nerfw} & 59.2 & 23.2 & 26.4 & 18.9  \\   \methodAblative & 64.4 & 23.8 & 26.8 & 19.6 \\
    \method (\textbf{ours}) & 66.7 & 24.0 & 27.2 & 19.8 \\
    \methodActor (\textbf{ours}) & \textbf{69.1} & 24.1 & \textbf{27.3} & 19.9 \\
    \methodColor (\textbf{ours}) & 67.4 & {24.1} & {27.2} & {19.9} \\
    \methodColorActor (\textbf{ours}) & 67.8 & \textbf{24.2} & \textbf{27.3} & \textbf{20.0} \\
    \bottomrule
  \end{tabular}
    \caption{\textbf{\benchmarkname.} Average mAP scores, which evaluate foreground segmentation, and average PSNR scores for the full scene (PSNR), the background part ($\text{PSNR}_b$), and the foreground part ($\text{PSNR}_f$) according to the segmentation.}
  \label{tab:epic}
\end{table}


\subsection{Results}
\label{sec:exp_results}

We compare our method and the different baselines on the \benchmarkname benchmark presented in the previous section.

\paragraph{Quantitative results.}
Table~\ref{tab:epic} compares the different methods according to the four evaluation metrics of \benchmarkname. This allows to evaluate their capacity to discover and segment 3D objects but also to reconstruct dynamic scenes. We make the following observations.

First, all flavors of our approach largely outperform the existing NeRF and NeRF-W. They also outperform the naive \methodAblative which simply combines two NeRF models respectively modeling the foreground and the background.

Second, including temporal information as an input to neural rendering (either with a low-rank expansion of the trajectory states as in \method or using a two stream architecture that processes time transformed with positional encoding as in \methodAblative) proves to be essential for improving the performance over NeRF and NeRF-W. 

Third, we observe that each of the proposed improvements, \methodActor and \methodColor, outperform the vanilla version of our approach. The third stream for modeling the actor brings +2.4 mAP points to the segmentation task while the better color model brings +0.7 mAP points. They also slightly improve the frame reconstructions (PSNR scores).

Finally, we see that combining the two proposed improvements (\methodColorActor) further improve new-view synthesis but not the foreground segmentation quality.
Note however that the segmentation metric does not reflect the ability of the full model to separate foreground into objects and actor (since they are merged together in the annotated masks).
This ability is illustrated in~\Cref{fig:splash}. 


\def\im#1{
    \fontfamily{lmss}\selectfont
    \scriptsize
    {\includegraphics[width=0.13\linewidth]{#1}}
 }
 
\begin{figure*}[t!]
    \small
    \centering
    \setlength{\tabcolsep}{1pt}
    \def\arraystretch{0.50}
    \begin{tabular}{ccccccc}
        & GT + Mask & NeRF-W~\cite{martinbrualla2020nerfw} & \method & \methodActor & \methodColor & \methodColorActor \\
        \addlinespace[1pt]
        \parbox[t]{5mm}{\multirow{4}{*}{\rotatebox[origin=c]{90}{Best segmentation masks}}} &
        \im{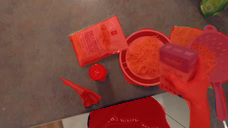}& \im{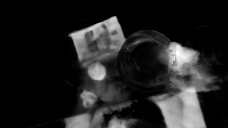}& \im{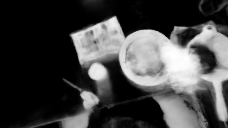}& \im{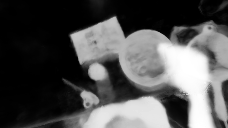}&
        \im{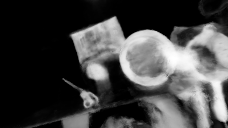}&
        \im{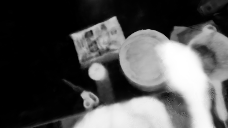}\\ & \im{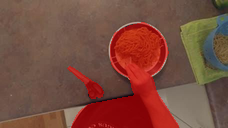}& \im{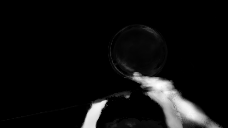}& \im{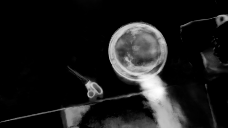}& \im{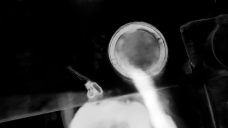}&
        \im{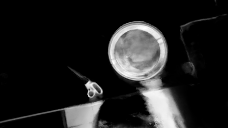}&
        \im{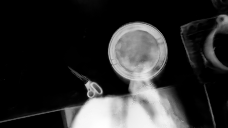}\\ & \im{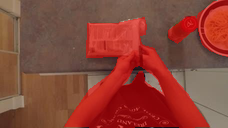}& \im{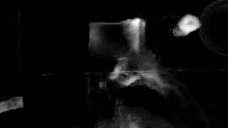}& \im{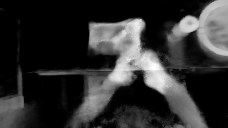}& \im{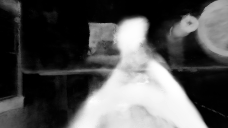}&
        \im{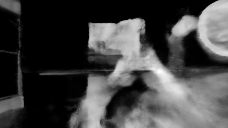}&
        \im{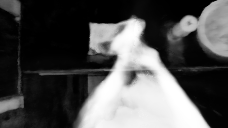}\\ & \im{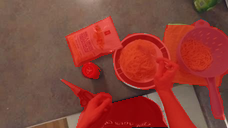}& \im{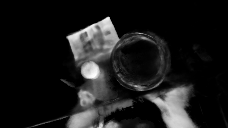}& \im{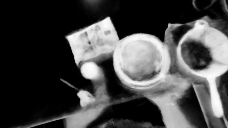}& \im{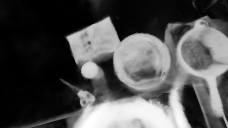}&
        \im{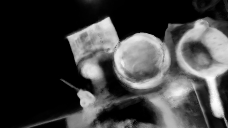}&
        \im{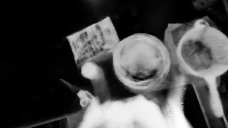}\\
        \parbox[t]{5mm}{\raisebox{-0.4\normalbaselineskip}[0pt][0pt]{\rotatebox[origin=c]{90}{Failure Cases}}} &
        \im{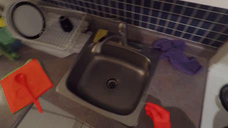}& \im{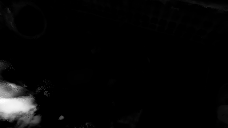}& \im{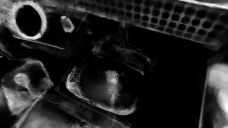}& \im{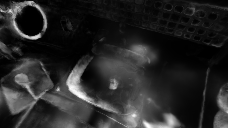}&
        \im{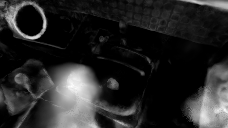}&
        \im{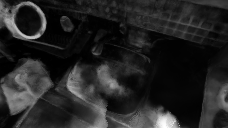} \\ & \im{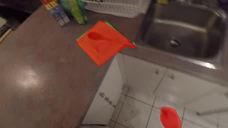}& \im{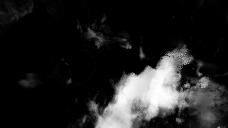}& \im{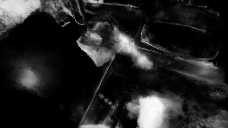}& \im{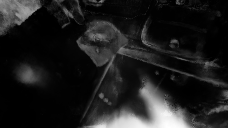}&
        \im{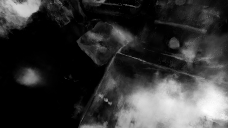}&
        \im{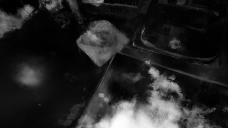}\\
    \end{tabular}
    \vspace{-0.3cm}
    \caption{\textbf{Segmentation masks.} For a given sequence, segmentation results of the 5 (resp. 3) frames for which \methodActor got the highest (resp. lowest) AP, for NeRF-W~\cite{martinbrualla2020nerfw}, \method, \methodColor, and \methodColorActor.}
    \label{fig:best5masks_2col}
    \vspace{-0.5cm}
\end{figure*}


\paragraph{Qualitative results.}
First, we verify the quality of the views reconstructed by our method.
\Cref{fig:quali_comparisons} compares three frames from three different scenes, with their reconstruction by NeRF, NeRF-W, \method and \methodColorActor.
As already observed~\cite{park2020nerfies,li21neural}, NeRF struggles with the dynamic components in the scene and produces blurry reconstructions.
NeRF-W obtains sharper reconstructions of the static regions, but does not handle well the dynamic regions. 
\method produces sharper results, especially for the moving objects. See for example the plates.
Finally \methodColorActor captures more details, such as the arms in the first scene, or the spoon in the second.

Next, \Cref{fig:best5masks_2col} illustrate success and failure cases for the segmentation task.
In the best cases for our method (top), NeRF-W produces noisy predictions which barely capture the plate, the pasta colander, and the actor's body.
\method improves over these results and captures all detachable objects and more body parts, but classifies part of the floor as foreground.
\methodActor successfully identifies the floor as static and better predicts the shape of the actor's body (second and third rows).
For the failure cases (bottom), while the segmentation mAP scores are generally low, \method outperforms NeRF-W even more significantly.
Specifically, NeRF-W fails to identify foreground objects.
\method improves over that but incorrectly classifies some of the background as foreground (see \eg rows 6 to 8 where the drying rack, static in this scene, is predicted as dynamic).
Such errors are reduced by \methodColor, \methodActor and \methodColorActor.

\paragraph{Comparison to motion segmentation.}

\begin{figure}[t!]
    \footnotesize
        \begin{tabular}{cccc}
             Frame + GT ~~ & NeuralDiff+A & ~~~ MG \cite{Yang21a} ~~~ & ~~~ MoSeg \cite{OB14b} ~~~ \\  
       \end{tabular}
           \includegraphics[width=0.24\linewidth]{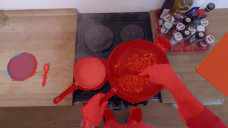}         \includegraphics[width=0.24\linewidth]{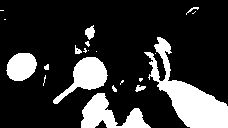}          \includegraphics[width=0.24\linewidth]{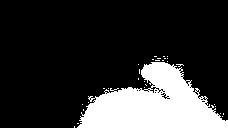}         \includegraphics[width=0.24\linewidth]{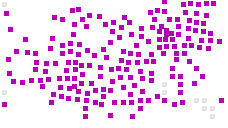} \\ 
           \includegraphics[width=0.24\linewidth]{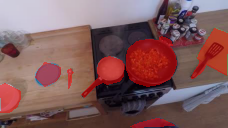}         \includegraphics[width=0.24\linewidth]{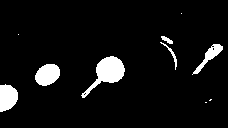}         \includegraphics[width=0.24\linewidth]{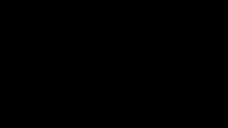}          \includegraphics[width=0.24\linewidth]{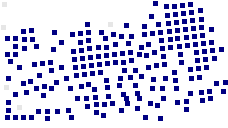} 
           \vspace{-0.5em}
    \caption{\textbf{Qualitative results.} Our method captures the body parts, even when currently not moving, and foreground objects, even if temporarily idle. MotionGroup (MG) \cite{Yang21a} only captures the actor when moving. The traditional MoSeg \cite{OB14b} is not designed for this challenging egocentric scenario and fails entirely.}
    \label{fig:quali_motionseg}
    \vspace{-0.2cm}
\end{figure}

\begin{table}[t!]
  \centering
  \small
  \begin{tabular}{lrrrrr}
    \toprule
    Method & & & & &IoU \\
    \midrule
    \textbf{NeuralDiff + A (ours)} & & ~~~~~ & & & \textbf{{43.1}} \\\relax
    MotionGroup \cite{Yang21a}$^{*} $ & & & & & {19.8} \\
    MotionGroup \cite{Yang21a}$^{**}$ & & & & &{28.7} \\
    \bottomrule
  \end{tabular}\vspace{-0.5em}
    \caption{Comparison with MotionGroup~\cite{Yang21a}. Average of per-frame IoU scores for all 10 sequences. $^{*}$ indicates that we use their pretrained model, and $^{**}$ that we retrained their model on one of our scenes.}
  \label{tab:motionseg}
\end{table}

Finally, we compare to MoSeg \cite{OB14b} as a traditional motion segmentation method and MotionGroup \cite{Yang21a} as a recent one. We report intersection over union (IoU) scores, as these methods produce binary masks. Quantitative and qualitative results are shown in \Cref{tab:motionseg} and \Cref{fig:quali_motionseg}. We see that none of these approaches is suited for the task that our method is designed for, as these techniques are unable to capture objects that move infrequently and suffer from the heavy occlusions of the actor. The method of \cite{OB14b} is particularly ill-suited for the task, so we did not evaluate it quantitatively.

\section{Conclusions}\label{s:conclusions}

We have shown that recent neural rendering techniques can successfully be applied  to the \emph{analysis} of egocentric videos.
We have done so by introducing \method, a triple-stream neural renderer separating the background, foreground and actor via appropriate inductive biases.
Together with other improvements such as smooth dynamics and more principled color mixing, \method significantly outperforms baselines such as NeRF-W for our task.

We have used \method to identify objects that move (\ie are detached) in long and complex video sequences from EPIC-KITCHENS, even if the objects are small and move only little and sporadically.
We believe that these results will inspire further research on the use of neural rendering for unsupervised image understanding.

\noindent\textbf{Acknowledgments.} Andrea Vedaldi was partially sponsored by ERC 638009-IDIU.

\bibliographystyle{ieee_fullname}
{\small\bibliography{biblio}}

\newpage
\appendix

\section*{Supplemental Material}

\section{Proof of equation (4)}

Section~\ref{sec:method} describes the problem of mixing colors from different model components, and introduces a more principled color mixing model to resolve this issue. This model is obtained by decomposing the segment $\delta_k$ in $Pn$ sub-segments, alternating between the $P$ different materials ($P=|\mathcal{S}|$, \eg $P = 3$ if the background, foreground and actor are considered).

In the limit, the probability that the photon is absorbed in a subsegment of material $p$ is given by:
\begin{equation}\label{e:mixsuppl}
w^p(T_k)
= \frac{\sigma^p_k}{\sum_{q=1}^P \sigma^q_k}\left(1 - \prod_{q=1}^P T^q_k\right).
\end{equation}
and we propose a proof for this claim below.

\begin{proof}[Proof of~\Cref{e:mixsuppl}]
Decompose segment $\delta_k$ in $Pn$ subsegments, alternating between materials $p \in \{1,\dots,P\}$ in a cyclic fashion.
The probability that material $p=1$ is responsible for the absorption is given by:
$$
\sum_{i=0}^{n-1} (1 - (T_k)^{\frac{1}{n}})^i (1-(T_k^p)^{\frac{1}{n}})
=
\frac
{1-(T_k^p)^{\frac{1}{n}}}
{1-\bar T_k^{\frac{1}{n}}}
(1-T_k),
$$
where
$
\bar T_k = \prod_{q=1}^P T_k^q
$
and
$
T_k^p = e^{-\sigma^p_k\delta_k}.
$
In the limit for $n\rightarrow \infty$, this expression reduces to
$
({\ln T_k^p}/{\ln \bar T_k}) (1-T_k)
$
which is the same as~\Cref{e:mix2}.
\end{proof}


\section{Implementation details}

\Cref{sec:exp_protocol} contains some implementation details about the
architecture. We provide further details below.

\paragraph{Architecture.} As outlined in \Cref{sec:method}, we make use of a three stream architecture to separate the background, foreground and actor.
Similar to \cite{martinbrualla2020nerfw}, we implemented $\MLP^b$ and $\MLP^f$ such that they share the weights of their initial layers. Let us define the set of shared
layers as $\MLP^s$. Given a ray $r_k$, the shared MLP encodes the ray as
$\rho_k$ and produces the background density with $(\rho_k, \sigma_k^b) =
\MLP^s(g_t r_k$). Same as the static model in \cite{martinbrualla2020nerfw}, $\MLP^b$ further processes $\rho_k$ and outputs the corresponding background color $(c_k^b) = \MLP^b(\rho_k, d_t, y_k^f)$ where $d_t = q_t / \|q_t\|_2$ with $q_t = g_t K^{-1} u$ is the respective unit normalized viewing direction and $y_k^f$ is the frame specific appearance code (equivalent to the latent appearance embedding in NeRF-W\footnote{While the EPIC-KITCHENS dataset has less variability in terms of photometric variation than the unconstrained photo collections used in NeRF-W~\cite{martinbrualla2020nerfw}, we observed that encoding appearance still results in better reconstructions.}). The foreground MLP takes the encoded ray and the frame specific code $z_t^f$ as input and produces the density, color and uncertainty score as in $(\sigma^f_k,c^f_k,\beta^f_k) = \MLP^f(\rho_k, z^f_t)$.

The actor MLP does not rely on $\MLP^s$, meaning it does not share weights with $\MLP^b$ and $\MLP^f$. Similar to $\MLP^f$ it also uses the frame specific code as input. It takes a ray that is relative to the camera, and the frame code as input and produces, analogously to $\MLP^f$, the density, color and uncertainty score with $(\sigma^a_k,c^a_k,\beta^a_k) = \MLP^a(r_k, z^a_t)$ with $z^a_t = z^f_t$. Same as in NeRF-W, we add a minimum importance $\beta_\text{min}$ (as hyperparameter) to the sum of the pseudo-standard deviations, resulting in $\beta_{ut} = \sum_{p}\beta^p_{ut} + \beta_\text{min}$, where $p$ is the material, and $u$ a pixel from frame $t$.

The layers of the shared MLP, $\MLP^s$, consists of 256 units, the other MLPs consists of 128 units. We respectively use 8, 1, 4, and 4 layers for $\MLP^s$, $\MLP^b$, $\MLP^f$, and $\MLP^a$.

\paragraph{Sampling points efficiently.} 
Analogously to NeRF~\cite{mildenhall20nerf:} and NeRF-W~\cite{martinbrualla2020nerfw}, we improve the sampling as described in \Cref{sec:basic_nerf} by simultaneously optimizing two volumetric radiance fields, a coarse one, $f^\text{coarse}$, and a fine one $f^\text{fine}$. Using both models enables us to sample free space and occluded regions that do not contribute to the rendered image less frequently by ``filtering'' these regions out with the coarse network. We achieve this by using the learned density of the coarse model to bias the sampling of the points along a ray for the fine model. Similarly to \cite{martinbrualla2020nerfw}, we apply the proposed architectural extensions (such as the actor model) only to the fine model. Therefore, in practice the training loss in \Cref{sec:uncertainty} (which describes the training of the fine radiance field $f^\text{fine}$) is extended with
\begin{equation*}
\mathcal{L^\text{coarse}}(f^\text{coarse}|x_t,g_t,u) = \sum_{u}{\| x_{ut} - f_u^\text{coarse}(g_t) \|^2},
\end{equation*}
where pixel $u \in \Omega = \{0,\dots,H-1\}\times \{0,\dots,W-1\}$. Our final loss is then $\mathcal{L} = \mathcal{L}_\text{prob} + \lambda \mathcal{L}_\text{sparse} + \mathcal{L^\text{coarse}}$.

\paragraph{Similarities and differences with NeRF and NeRF-W.} The rendering mechanism is virtually the same as the one described in NeRF \cite{mildenhall20nerf:} with the exception that we use a batch size of 1048 rays, and sample 64 points along each ray in the coarse volume and 64 additional points in the fine volume. Similarly to \cite{martinbrualla2020nerfw}, we set $\beta_\text{min} = 0.03$, and apply positional encoding on the inputs. In comparison, we use $256$ units for the shared MLP (referred to as $\MLP_{{\theta}_1}$ in \cite[p.4]{martinbrualla2020nerfw}) and do not omit the color and density from the foreground model (see \cite[p.5]{martinbrualla2020nerfw}), as we use it for the final rendering.

\paragraph{Training.} All models are trained separately for each scene for 10 epochs on 1 GPU, taking approximately 24 hours with an NVIDIA Tesla P40. We downscale the images extracted from the videos of the EPIC Kitchens dataset to a resolution of $128 \times 228$. The different hyper-parameters are selected on the validation set via grid search to improve the photometric reconstruction.

\begin{table}[t!]
  \centering
  \begin{tabular}{ccccccc}
    \toprule
    ID & KID & Train & Val. & Test & $\text{Ann.}$ & Duration \\
    \midrule
    01 & P01\_01 & 752 & 54 & 54 & 54 & 27 min \\
    02 & P03\_04 & 794 & 57 & 57 & 56 & 28 min \\
    03 & P04\_01 & 797 & 57 & 57 & 57 & 19 min \\
    04 & P05\_01 & 808 & 58 & 58 & 58 & 06 min \\
    05 & P06\_03 & 867 & 62 & 62 & 61 & 11 min \\
    06 & P08\_01 & 656 & 47 & 47 & 47 & 10 min \\
    07 & P09\_02 & 757 & 54 & 55 & 54 & 06 min \\
    08 & P13\_03 & 689 & 49 & 50 & 50 & 06 min \\
    09 & P16\_01 & 838 & 60 & 60 & 60 & 20 min \\
    10 & P21\_01 & 867 & 62 & 62 & 61 & 11 min \\
    \midrule
    - & All & 7825 & 560 & 562 & 558 & 144 min \\
    \bottomrule
  \end{tabular}
  \caption{\textbf{Summary of \benchmarkname.} Number of frames per scene for training, evaluation, and testing. Including annotations for test frames. KID refers to the video ID from the EPIC Kitchens dataset. Note that some frames do not show any detachable objects or the actor, hence resulting in some cases in fewer annotations than test set frames.}\label{tab:summary_scene_epicdiff}
\end{table}

\paragraph{Testing.}
The weights of a model used for representing one complete scene take about 17MB of disk space. We render the views of one entire scene in about one hour with an NVIDIA GeForce RTX 2080.

\section{Details about the \benchmarkname benchmark}

\begin{figure}[t!]
    \centering
        \includegraphics[width=0.48\linewidth]{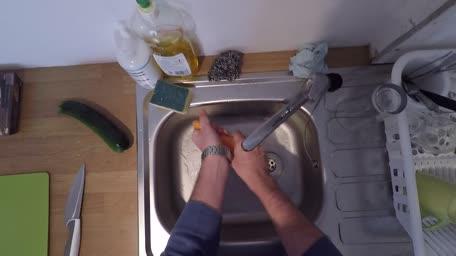} 
        \includegraphics[width=0.48\linewidth]{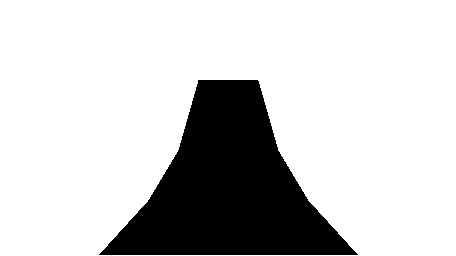} 
    \caption{\textbf{Image and mask as input for COLMAP.} The mask  approximates the location of the hands over all frames. COLMAP will use the features found in the white area and will ignore the masked ones (black).}
    \label{fig:colmap_mask}
\end{figure}

\begin{figure}[t!]
    \centering
        \includegraphics[width=1.\linewidth]{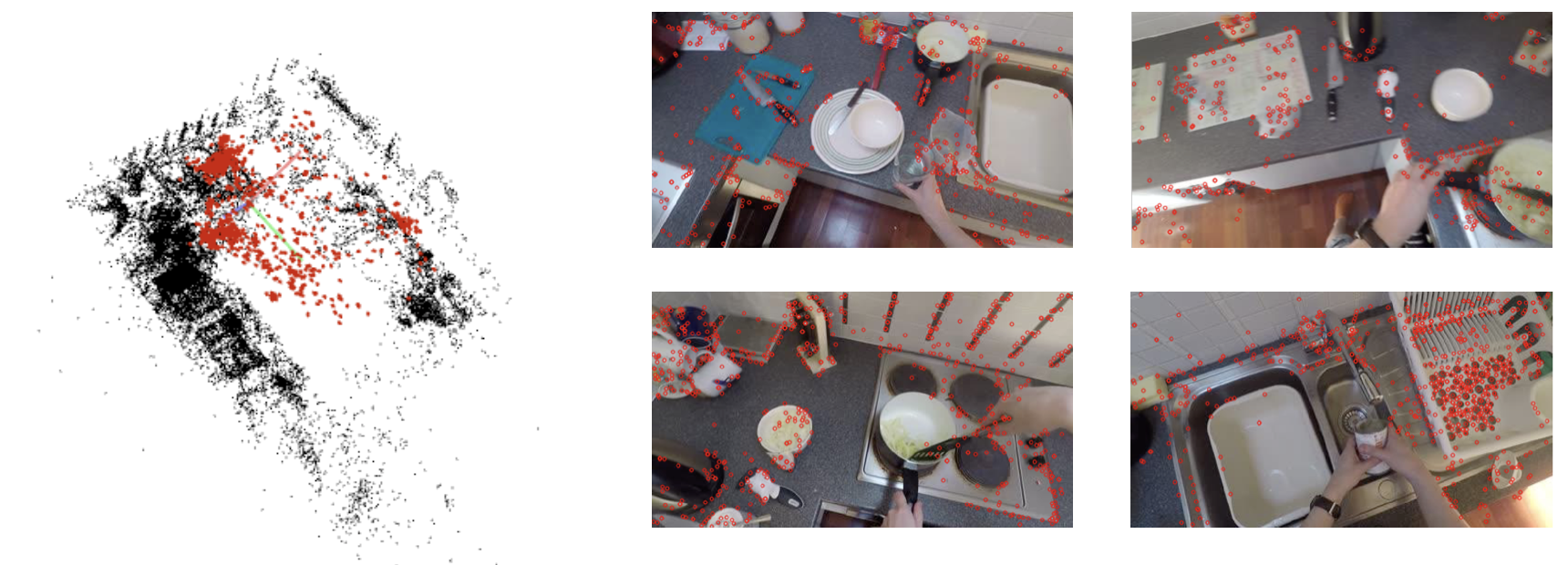}
        \includegraphics[width=1.\linewidth]{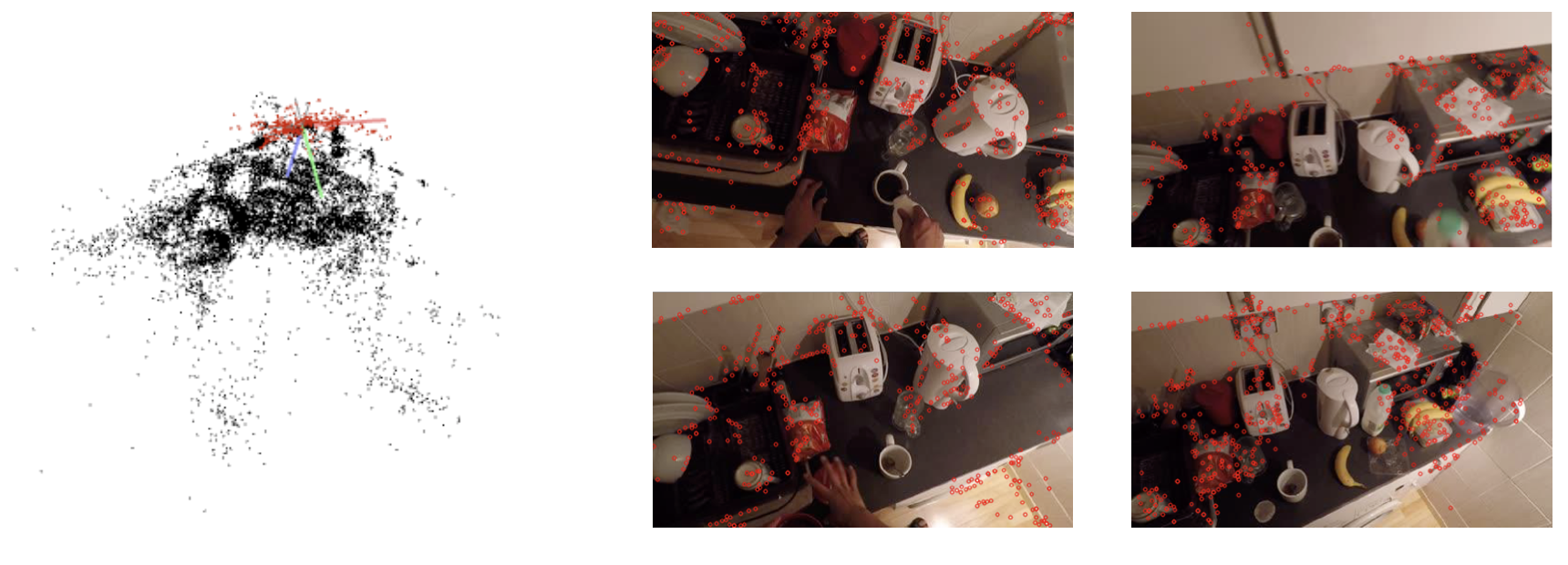}
        \includegraphics[width=1.\linewidth]{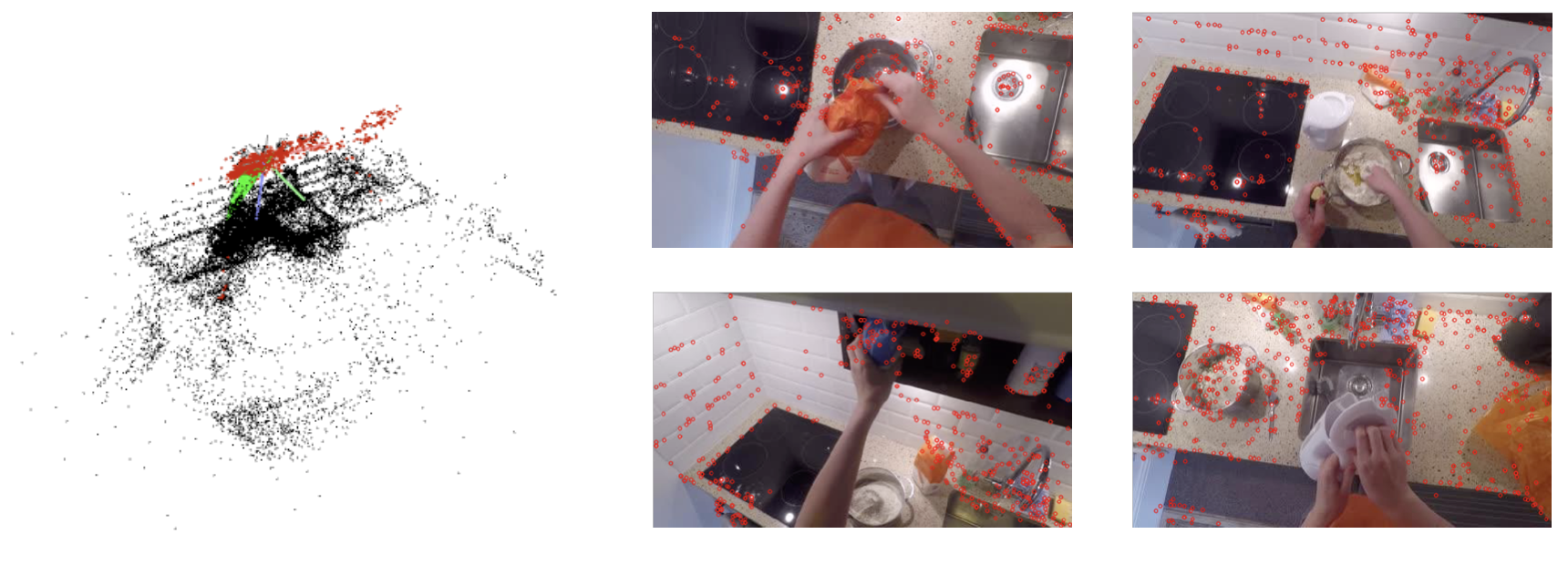}
        \includegraphics[width=1.\linewidth]{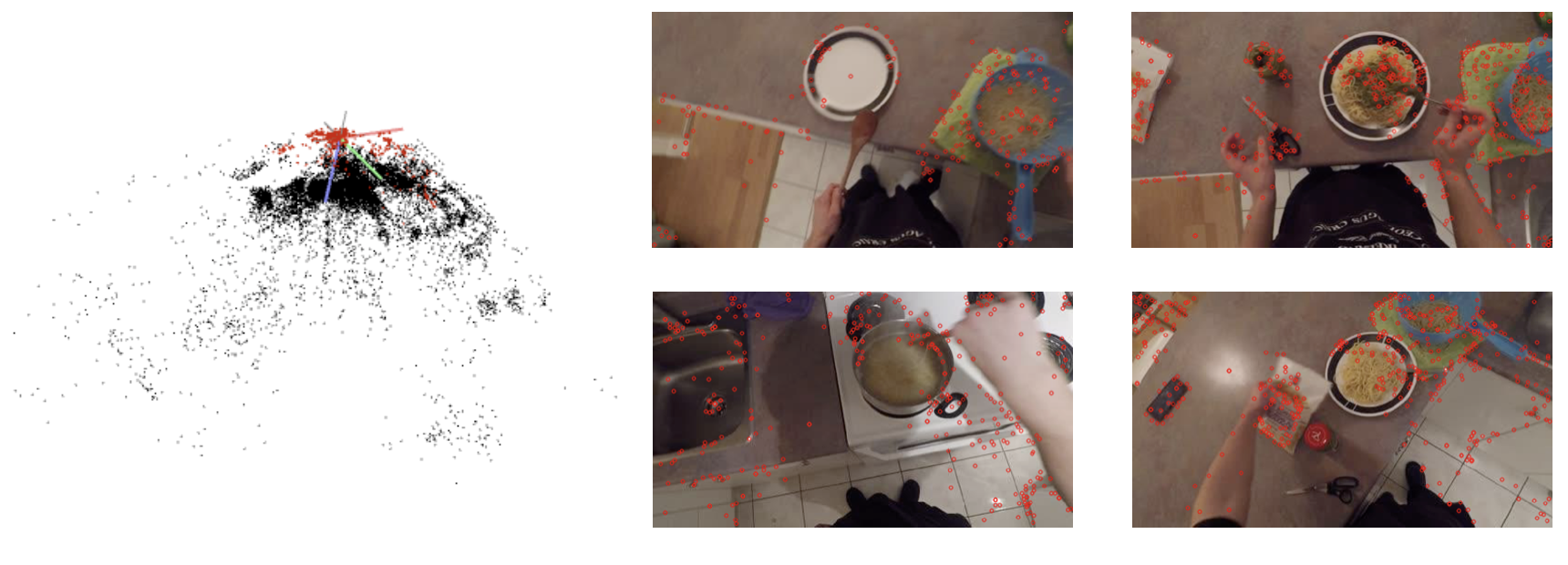}
        \includegraphics[width=1.\linewidth]{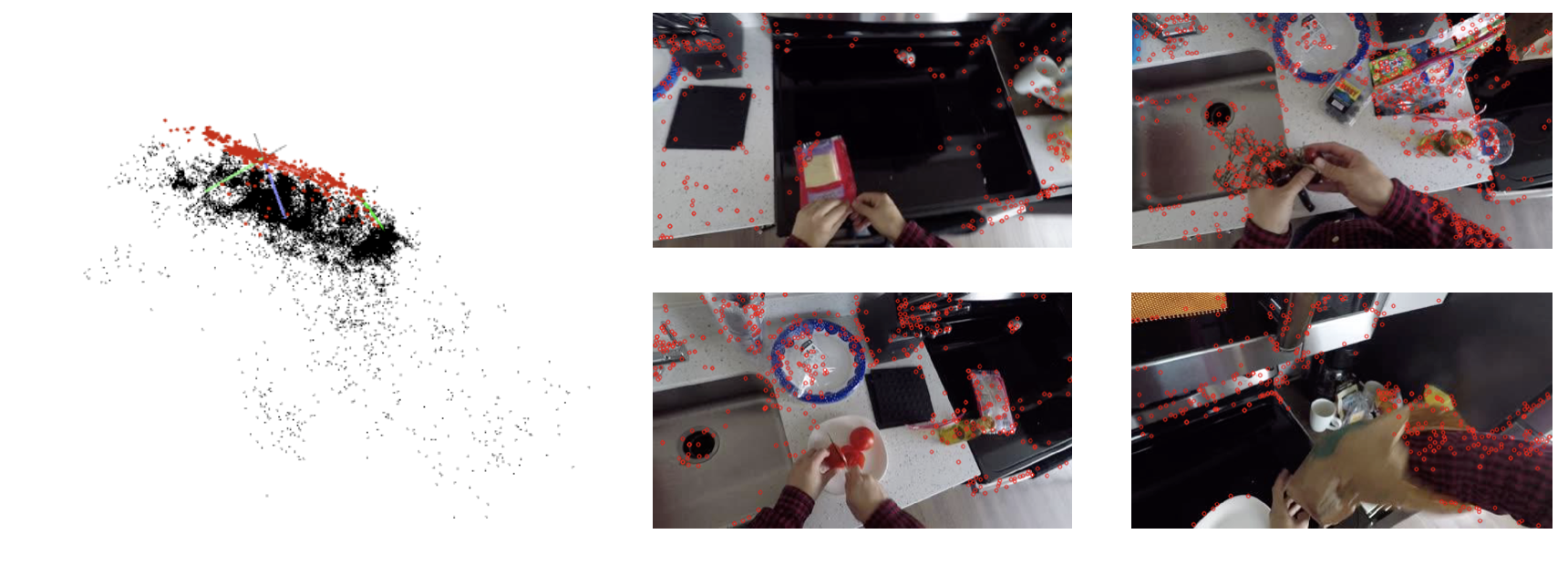}
    \caption{\textbf{Sparse reconstructions of 5 scenes.} We visualize point clouds for 5 scenes, with example images showing different views of each scene with their corresponding extracted features. The red dots in the point clouds represent estimated extrinsic camera parameters. Note that the bottom middle part of each image does not contain features as explained in \Cref{fig:colmap_mask}.}
    \label{fig:pointclouds}
\end{figure}

This section extends \Cref{s:dataset} from the main paper. We give additional details about the way we created the \benchmarkname benchmark out of the EPIC-KITCHENS dataset~\cite{Damen2020RESCALING}.

As a first pre-filtering step, we ignored videos where the person is mostly washing dishes and we selected through visual inspection the scenes that show a high variety of viewpoints and manipulated objects. We then extracted the frames from these videos and sampled each $i$-th frame (linearly), where $i$ is chosen such that we have 1000 frames per scene. In the next step we used COLMAP's feature extractor with SIFT to calculate keypoints in the most reliable frame regions (\ie, we excluded regions that are likely to correspond to the actor and not the scene, using a fixed mask, as shown in \Cref{fig:colmap_mask}). Then we match the features with vocabulary feature matching, and create a sparse 3D reconstruction. This results in a subset of the 1000 frames that get registered. We further filter the scenes by constraining them to have a minimum of 600 registered frames (out of the 1000 initial ones). We split the remaining frames of each scene into a train, a validation and a test set, where we select each $16$-th frame for validation and every other $16$-th frame for testing. The remaining frames are then used for training. 

For the foreground segmentation task, we annotate all images from the test set with the VGG Image Annotator \cite{dutta2016via, dutta2019vgg} for 10 scenes from the EPIC-KITCHENS dataset \cite{Damen2020RESCALING}, filtered with the procedure described above. For this task, we define the foreground as all moving objects \textit{and} the actor. A summary of the dataset statistics can be found in \Cref{tab:summary_scene_epicdiff}. This table shows the number of frames per train, validation and test set with the latter's corresponding to annotated frames. For \benchmarkname, we extracted a total number of about 9000 frames from about 140 minutes of video material for the 10 scenes, and annotated 558 frames out of these frames. Sparse reconstructions for 5 out of the 10 scenes can be found in \Cref{fig:pointclouds}.

\section{Segmentation precision-recall curves}

\begin{figure}
\begin{center}
 \includegraphics[width=0.95\linewidth]{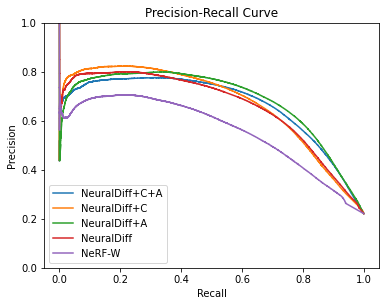} 
\end{center}
   \caption{\textbf{Precision-Recall Curve calculated over all scenes.} We combine the predicted and target masks from all scenes and calculate the average precision and recall over all the pixels.}
\label{fig:pre_rec_curve}
\end{figure}

We evaluate the capacity of the methods to discover and segment 3D objects with a precision-recall curve in \Cref{fig:pre_rec_curve}. 
We calculate each curve by taking the prediction scores and ground truth masks for all the pixels of all test frames from all scenes, and then calculating the precision and recall with varying thresholds for the prediction scores.
This leads to observations similar to the ones we made for \Cref{tab:epic} in \Cref{sec:exp_results}. More precisely, we observe that, for any recall, NeRF-W exhibits a lower precision than any flavor of our approach.  
We can also see that \method has a lower precision than \methodActor, \methodColor, and \methodColorActor, indicating that \method benefits from the actor model and color normalization (individually and combined).

\end{document}